\documentclass{article}

\usepackage{microtype}
\usepackage{graphicx}
\usepackage{subfigure}
\usepackage{booktabs} 

\usepackage{hyperref}

\usepackage[accepted]{icml2024}

\usepackage{amsmath}
\usepackage{amssymb}
\usepackage{mathtools}
\usepackage{amsthm}

\usepackage[capitalize,noabbrev]{cleveref}

\theoremstyle{plain}

\theoremstyle{definition}

\theoremstyle{remark}

\usepackage[textsize=tiny]{todonotes}

\icmltitlerunning{Understanding the Cognitive Complexity in Language Elicited by Product Images}

\begin{document}

\twocolumn[
\icmltitle{Understanding the Cognitive Complexity in Language Elicited by Product Images}



\icmlsetsymbol{equal}{*}

\begin{icmlauthorlist}
\icmlauthor{Yan-Ying Chen}{equal,comp}
\icmlauthor{Shabnam Hakimi}{equal,comp}
\icmlauthor{Monica Van}{equal,comp}
\icmlauthor{Francine Chen}{comp}
\icmlauthor{Matthew Hong}{comp}
\icmlauthor{Matt Klenk}{comp}
\icmlauthor{Charlene Wu}{comp}
\end{icmlauthorlist}

\icmlaffiliation{comp}{Toyota Research Institute, Los Altos, California}

\icmlcorrespondingauthor{Yan-Ying Chen}{yan-ying.chen@tri.global}

\icmlkeywords{Machine Learning, ICML}

\vskip 0.3in
]



\printAffiliationsAndNotice{\icmlEqualContribution} 

\begin{abstract}
Product images (e.g., a phone) can be used to elicit a diverse set of consumer-reported features expressed through language, including surface-level perceptual attributes (e.g., ``white'') and more complex ones, like perceived utility (e.g., ``battery''). The \textit{cognitive complexity} of elicited language reveals the nature of cognitive processes and the context required to understand them; cognitive complexity also predicts consumers' subsequent choices. This work offers an approach for measuring and validating the cognitive complexity of human language elicited by product images, providing a tool for understanding the cognitive processes of human as well as virtual respondents simulated by Large Language Models (LLMs). We also introduce a large dataset that includes diverse descriptive labels for product images, including human-rated complexity. We demonstrate  that human-rated cognitive complexity can be approximated using a set of natural language models that, combined, roughly capture the complexity construct. Moreover, this approach is minimally supervised and scalable, even in use cases with limited human assessment of complexity. \footnote{The data will be available at \url{https://github.com/TRI-MAC/cognitive_complexity}}

\end{abstract}

\section{Introduction}
Language can reveal how \textit{context} shapes the way consumers evaluate products, signaling the range of psychological processes underlying their assessments. Previous work has shown how models of language capture stimulus feature processing using properties like concreteness, or the degree to which a word refers to a perceptible entity in mind \cite{Brysbaert2014,charbonnier-wartena-2019-predicting}, while other models use language to infer a person's psychological state by quantifying factors such as sentiment \cite{zhang-qian-2020-convolution}, emotion \cite{demszky-etal-2020-goemotions}, and personality \cite{flekova-gurevych-2015-personality}.  
Such models have not yet elucidated how the complexity of thought processes influences choice. We examine cognitive complexity, or the \textit{additional context needed outside of an eliciting visual stimulus} to understand or explain human-generated language \cite{hakimi2023}, as a means for inferring the relationship between thoughts and actions. For example, when viewing a specific product image, such as a red car, some might focus on the dominant perceptible feature---the color red---while others may comment on more seemingly ``complex'' features shaped both by perception and projected utility and experiences, like ``engine.'' Both words are concrete but involve different levels of complexity in processing visual stimuli.

In this light, a measure of cognitive complexity should make individual variability in language generation and associated underlying preferences more interpretable because it quantifies a shared, underlying structure between idiosyncratic representations. For example, given the same product image stimulus, two people might describe very different evoked memories; although the contents of these descriptions may differ greatly in other natural language measures, both sets of text would have high complexity. By leveraging shared features of the processes that generated these descriptions, we can more accurately and effectively predict subsequent behaviors (e.g., choices, language use) both for the two individuals and the population as a whole. Indeed, previous work demonstrated that cognitive complexity explains unique variance in preferences and improves choice prediction \cite{hakimi2023}. Because cognitive complexity captures the process through which a stimulus can potentiate emergent psychological phenomena---i.e., thoughts and behaviors that are not intrinsic to the stimulus itself---it can also be used to infer whether the language was generated by a human or an LLM imitating a human. For example, an observed discrepancy between human- and LLM-generated responses in the distribution of cognitive complexity scores for certain visual stimuli can be used to filter out unqualified subjects or data \cite{veselovsky2023prevalence}, e.g., subjects who submit machine-generated responses to a survey that requires human responses. Similarly, distributional alignment can also be used to evaluate an LLM's performance in generating language with human-like cognitive complexity.

Here, we validate a computational measure of cognitive complexity by interrogating its relationship with human ratings as well as related natural language constructs. Because cognitive complexity is a multidimensional construct \cite{ConwayIII2014,glaze2018}, we evaluate the putative informational content of the ``additional context" needed for understanding by testing the relationship between cognitive complexity and four related constructs---visibility, semantics, uniqueness, and concreteness---that have been reliably encoded using validated vision and language models. Further, to verify how well cognitive complexity aligns with human judgments, we first elicit user-generated text labels for product images (as \cite{hakimi2023}) then compare computed to human-rated measures. This work presents the first large and diverse dataset collected to understand the cognitive complexity of product image descriptions. 

Our contributions include: (1) the first dataset with 4000+ product images across 14 categories and 45,609 human-generated text labels and complexity ratings for computing cognitive complexity, and (2) the use of relevant vision and language models to characterize and approximate the cognitive complexity of language elicited by product images.

\section{Related Work}
Images provide information that shapes the cognitive process \cite{cavanagh2011visual, marr2010vision}. Hence, what is visible to people may influence the cognitive complexity of their reactions. Recent work on joint text and vision embeddings has emphasized the strong ties between vision and language semantics for cognitive understanding in such diverse domains as hate speech recognition \cite{kiela2020} and noun property prediction \cite{yang-etal-2022-visualizing}. Such work may lead to improved models that exploit unimodal priors.  

Besides visibility and semantics, constructs built on language itself such as readability and uniqueness of words can influence text simplification \cite{maddela-xu-2018-word}. Concreteness of words that were rated by humans \cite{Brysbaert2014, iliev2016} or measured by learned word embeddings \cite{charbonnier-wartena-2019-predicting,wartena-2022-geometry} has been used to measure abstractness. \cite{hessel-etal-2018-quantifying} explored word concreteness in four image datasets by looking at the image clustering structure. \cite{ramakrishnan2021non} found that GPT word embeddings captured all information in brain imaging data for concreteness prediction based on logistic regression models.
However, the granularity of word concreteness is not grounded in a particular image. Cognitive complexity differs in that it assumes a joint vision-language representation: the complexity of a word is specific to the image it describes because reactions to different images vary. Moreover, cognitive complexity is a large theoretic construct \cite{ConwayIII2014} that overlaps with multiple constructs as discussed above.

\section{Data Elicitation} \label{sec:data}

\begin{figure}[tp]
    \centering
    \includegraphics[width=0.99\linewidth]{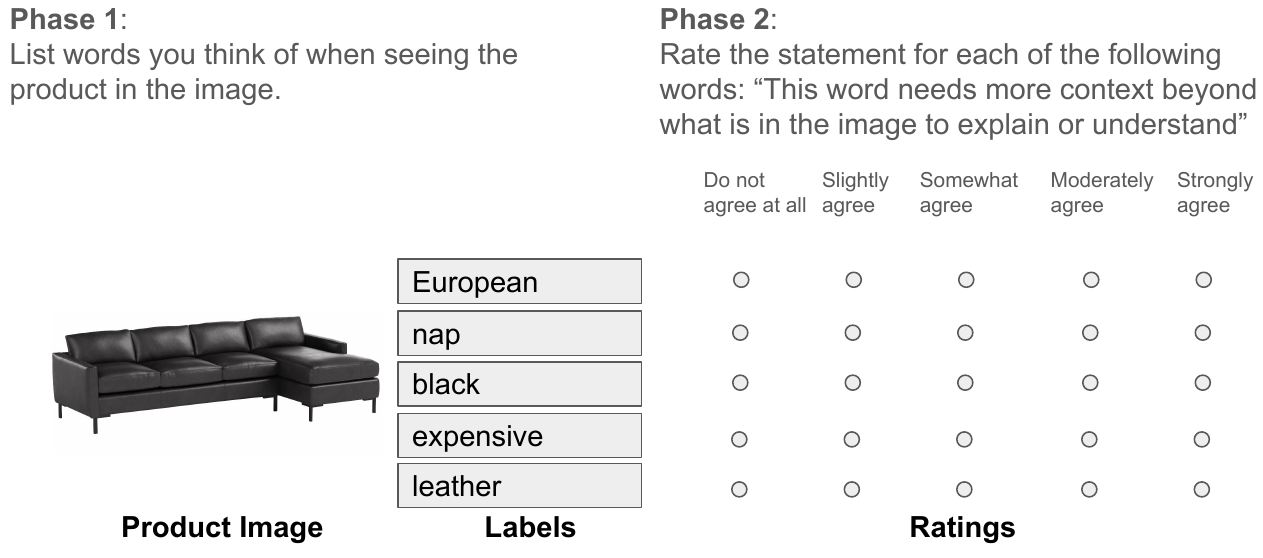}
    \caption{Data collection paradigm.}
    \label{fig:paradigm}
\end{figure}

\begin{table}
\centering
\begin{tabular}{c|cccc}
\hline
\textbf{Category} & \textbf{images} & \textbf{labels} & \textbf{vocab.} & \textbf{complexity}\\
\hline
furniture & 2,174 & 23,210 & 3,278 & 1.75 $\pm$0.94\\
decor & 1,054 & 10,481 & 2,664 & 1.77 $\pm$0.95\\
car & 181 & 2,925 & 741 & 1.90 $\pm$0.95\\
bed/bath & 239 & 2,686 & 871 & 1.76 $\pm$0.96\\
all & 4,093 & 45,609 & 6,583 & 1.77 $\pm$0.94
\\\hline
\end{tabular}
\caption{Dataset with product image and label counts as well as vocabulary size and complexity ratings (mean$\pm$sd) for the largest four product categories. 
See all 14 categories in Appendix Table \ref{tbl:data_full_cate}.}
\label{tbl:data}
\end{table}

In a series of studies, language was used to assess humans' representation of and preferences for a variety of real consumer products, either cars\footnote{Car images are licensed by EVOX \cite{evox}.} or items from the Amazon Berkeley Object (ABO) dataset \cite{hakimi2023,collins2022abo}.  The complexity of these responses to  product images was then evaluated by separate human raters. 

In Phase 1 (Fig. \ref{fig:paradigm}), 3,009 crowdsourced \cite{mturk} labelers were shown a product image and asked to ``list words you think of when seeing the product in the image.'' 
Each labeler was shown 16 images sampled from diverse product categories, and listed 5 to 20 text labels for each image. 
Note that the image background was subtracted and the images with text printed on products were removed to eliminate bias.
Overall 45,609 text labels for 4,093 images were collected. Data statistics are reported in Table \ref{tbl:data} and Appendix. 

As in prior work \cite{ConwayIII2014,Brysbaert2014}, 
human ratings are a reasonable proxy of ground truth representing human intuition and are used to verify subjective measures. Hence, 
in Phase 2 (Fig. \ref{fig:paradigm}, we asked a separate group of participants \cite{prolific} to rate the complexity of the previously elicited text labels, allowing for the estimation of ground truth that could be used to evaluate each model. 

Raters were not explicitly asked to rate the "complexity" of the words, since pilot results indicated that participants were confused by the ambiguity of determining whether a word was "complex" and often shifted how they interpreted the definition of "complex". Thus, raters were asked to instead determine whether product images provided sufficient context to interpret the generated words. 

For each product image and an associated set of 5-20 text labels, 1,210 raters assessed whether "this word needs more context beyond what is in the image to explain or understand" using a unipolar, 5-point Likert-style scale: "Do not agree at all" (0) to "Strongly agree" (4), which are converted to the complexity ratings (0-4). 
At least three (but not all) raters rated each label.  

\section{Approximating Cognitive Complexity} \label{sec:approximating}
Because cognitive complexity is a broad psychological construct that likely includes multiple sub-constructs, we triangulate it using the five models described below.
We hypothesized that these relevant, quantifiable constructs could be used together to approximate the complexity. 
We list hypotheses for each construct in italics.

\textbf{Visibility as Complexity}. 
Visual features that can be recognized at a glance, e.g., a primary color, may drive less complex cognitive processes while less visually salient features
may be more complex. 
We therefore leveraged text used to visually describe product images as a reference corpus to represent what people can intuitively see and recognize. 
A reference corpus for a product image is generated by a transformer-based image-to-text generation model OFA \cite{wang2022ofa}. It is trained by visual description datasets such as MSCOCO \cite{XinleiChen2015} and generates multiple caption-like outputs using beam search to expand the most possible words with variability. Because annotators for MSCOCO were asked to describe all the important parts of the scene, and not to describe things beyond the scene, the training datasets and approach force the generated descriptions to be visually perceived intuitively. 
We formulate the complexity score of a text label $l_m$ elicited by an image $I_n$, namely the complexity based on visibility (V) construct, as $\Theta_v(l_m,I_n)=1-f(l_m, D_n)$, where visibility $f(\cdot)$ means the number of sentences containing $l_m$ divided by the total number of sentences in the reference visual descriptions $D_n$ of $I_n$. Negating $f(\cdot)$ inverts its relationship with the score \textit{$\Theta_v$, i.e. the lower visibility, the higher complexity}.

\textbf{Semantics as Complexity.} Semantic intuition also influences complexity of elicited language such as the comprehension of language referring to the target product image, which is relevant to the semantic relationships between product images and elicited language.
We leverage a text and image embedding space based on CLIP \cite{pmlr-v139-radford21a} to measure the semantic relationship. Given a product image $I_n$ and a text label $l_m$, we encode the input to the joint embedding space and compute the cosine similarity. The complexity score is formulated as $\Theta_s(l_m,I_n)=1-\cos(E_T(l_m),E_X(I_n))$, where $E_T$ and $E_X$ are the text and image encoders, repectively. \textit{A text label is assigned with a higher complexity $\Theta_s$ if it is more semantically distant to the image stimulus (less similarity)}.


\textbf{Uniqueness as Complexity}. In addition to word features (e.g., length), we consider corpus-based word frequency \cite{maddela-xu-2018-word} to reflect word complexity. We sort the text labels into multiple corpora by the product categories of their corresponding product image stimulus. We then calculate the term frequency of a text label in its corpora. \textit{The complexity $\Theta_u$ of a text label is higher if it is less frequently used to describe product images of the same category, thus being more unique}.

\textbf{Concreteness as Complexity}. 
We use concreteness ratings collected by \cite{Brysbaert2014} for 40,000 English lemmas as a lookup table $B$ to derive the concreteness $c_m$ of each text label $l_m$ elicited by the product images. The complexity score is formulated as $\Theta_c(l_m)=b-c_m$, where b is set to the maximum score in $B$. The concreteness $c_m$ has an inverse relationship with $\Theta_c$, i.e., \textit{the lower concreteness, the higher complexity}.



\section{Experiments}
Our experiments are designed to investigate (1) how well the cognitive complexity approximated by the models aligns with human judgments and (2) whether the models are complementary to each other. In Table \ref{corr_human_whole}, we use the dataset introduced in Sec. \ref{sec:data} for evaluation. We also present results for a subset of this dataset in Table \ref{corr_human_high}, where product-label pairs obtained higher agreement (correlation $>$ 0.75) among multiple human raters, to reveal the influence of agreement on the results. 
Due to space limitations, the results for all 14 categories are reported in Appendix. 


\subsection{How Well Do the Models Align with Humans?}
While human ratings do not comprehensively represent the ground truth, they can be used to test the alignment of a model with human thoughts. Here, we calculate the Spearman correlation between the complexity rated by humans and as measured by models based on different constructs to verify our hypotheses in Sec. \ref{sec:approximating}. 

The results in Table \ref{corr_human_whole} and \ref{corr_human_high} suggest that models of visibility, semantics, uniqueness, and concreteness are correlated with human judgments as we hypothesize (hypotheses are in italics in Sec. \ref{sec:approximating})
The visibility and semantics constructs are top performers, perhaps because the two models use both text and image for the measurement, while other models measure complexity purely using a label or its metadata. This suggests that an image stimulus itself might provide meaningful information for computing cognitive complexity.

\begin{figure}[tp]
    \centering
    \includegraphics[width=0.99\linewidth]{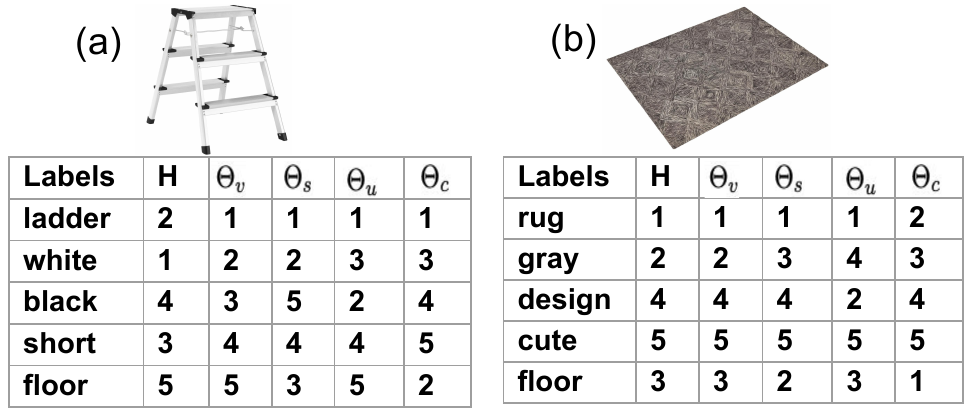}
    \caption{Ranking cognitive complexity (low to high) of text labels elicited by product images, rated by human (H) and measured by the models of individual constructs}
    \label{fig:example_complexity}
\end{figure}

Data also suggest that the alignment between each construct and human judgment varies across different product categories. For example, the visibility construct best aligns with humans for decor while semantics alignment is better aligned for furniture. As shown in Fig. \ref{fig:example_complexity}, the proposed constructs can help dive into different facets for efficient sense-making. A combination of multiple constructs may help generalize the measurement to various products. Note that significant individual differences  \cite{hakimi2023} suggest variance in the driving psychological factors supporting cognitive complexity. As such, high correlations with any sub-construct are unlikely.

\begin{table}
\centering
\begin{tabular}{c|cccccc}
\hline
\textbf{Models} & \textbf{furn.} & \textbf{deco.} & \textbf{car} & \textbf{bed} & \textbf{all}\\
\hline
 $\Theta_v$ & .230 & \textbf{.247} & \textbf{.281} & .239 & .225\\
 $\Theta_s$ & \textbf{.247} & .219 & .252 & \textbf{.247} & \textbf{.228}\\
$\Theta_u$ & .175 & .177 & .207 & .230 & .168 \\
$\Theta_c$ & .180 & .178 & .171 & .208& .160

\\\hline\hline
$\Theta_{v,s}$ & .258 & .252 & .266 & .259 & .250\\
$\Theta_{v,s,u}$ & .267 & .270 & .272 & .276 & .261\\
$\Theta_{v,s,u,c}$ & \textbf{.273} & \textbf{.276} & \textbf{.316} & \textbf{.294} & \textbf{.271}
\\\hline
\end{tabular}
\caption{Correlation with human ratings of complexity. Results for all 14 categories in Table \ref{corr_human_whole_all_cate_1} and \ref{corr_human_whole_all_cate_2} in Appendix.}
\label{corr_human_whole}
\end{table}

\begin{table}
\centering
\begin{tabular}{c|ccccc}
\hline
\textbf{Models} & \textbf{furn.} & \textbf{deco.} & \textbf{car} & \textbf{bed} & \textbf{all}\\
\hline
$\Theta_v$ & .391 & \textbf{.441} & \.471 & .513 & \textbf{.418}\\
$\Theta_s$ & \textbf{.425} & .375 & \textbf{.497} & \textbf{.520} & .414\\
$\Theta_u$ & .293 & .287 & .355 & .466 & .260 \\
$\Theta_c$ & .277 & .278 & .190 & .404 & .267 
\\\hline\hline
$\Theta_{v,s}$ & .436 & .443 & .509 & .550 & .433\\
$\Theta_{v,s,u}$ & \textbf{.459} & .458 & .531 & .611 & .456 \\
$\Theta_{v,s,u,c}$ & .454 & \textbf{.462} & \textbf{.543} & \textbf{.614} & \textbf{.472}

\\\hline
\end{tabular}
\caption{Correlation with human ratings of complexity (the subset with high agreement among raters). Results for all 14 categories in Table \ref{corr_human_high_all_cate_1} and \ref{corr_human_high_all_cate_2} in Appendix.}
\label{corr_human_high}
\end{table}

\subsection{Are the Models Complementary?}
We want to verify whether the models of constructs offer complementary information for measuring cognitive complexity. First, we compute the partial correlation between every two constructs while the rest of the constructs are the controlling variables. Table \ref{construct_corr} shows that some constructs have very low correlations such as uniqueness vs. concreteness. Visibility $\Theta_v$ that is measured based on what people visually perceived has higher correlations with other constructs as all these models consider perceptual information in different ways.
However, all pairs of constructs are far from identical and with each construct providing unique information.

The bottom part of Table \ref{corr_human_high} presents the correlations between combinations of top-performing models of constructs and human judgments. A combination takes the weighted sum of scores predicted by each construct, where the weight is empirically determined by their individual correlation with human judgment. In most cases, the listed combinations align with humans better than individual constructs. Adding more positively correlated constructs in combination usually increases the correlation with human judgments, probably because parts of the non-overlapped information are complementary. The findings are consistent with experiments based on the whole dataset in Table \ref{corr_human_whole}.


\begin{table}
\centering
\begin{tabular}{c|cccc}
\textbf{Constructs} & \textbf{$\Theta_v$} & \textbf{$\Theta_s$} & \textbf{$\Theta_u$} & \textbf{$\Theta_c$}\\\hline
\textbf{visibility ($\Theta_v$)} & 1 & - & - & -\\
\textbf{semantics ($\Theta_s$)} & .343 & 1 & - & -\\
\textbf{uniqueness ($\Theta_u$)} & .352 & .070 & 1 & -\\
\textbf{concreteness ($\Theta_c$)} & .283 & .168 & .040 & 1
\\\hline
\end{tabular}
\caption{Partial Correlation between the models  for measuring complexity based on different constructs}
\label{construct_corr}
\end{table}

\section{Discussion and Conclusion}
Human cognition is complex and diverse. While LLMs have reached or surpassed human in several knowledge-based tasks, there remain many challenges to approximating the kind of human cognitive complexity that characterizes subjective experience and behaviors such as elicited responses to products. As a further step to triangulate  complexity computationally, we propose grounding cognitive complexity in the language elicited by product images and characterize the constructs that might be predictive of this new measure. The data elicitation for cognitive complexity is more complicated than typical image annotations. First, it requires sampling a diverse population of participants, in this case, representing consumers. Second, it increases the labeling effort from the size of language data $O(N)$ to the size of pairs of language and image data $O(N \times M$). Hence, we foresee the cost of obtaining huge training data is nontrivial and attempt to investigate whether the foundational models, e.g., OFA \cite{wang2022ofa}, CLIP \cite{pmlr-v139-radford21a}, help compute cognitive complexity and generalize the measure to domains that are essentially different from the presented elicitation framework. Future work includes fine-tuning strategies to optimize the combination of constructs for better prediction power and broader use of measuring cognitive complexity in feedback to consumer products or other visual stimuli, and in responses synthesized by LLMs.


\nocite{langley00}

\bibliography{cogcom}
\bibliographystyle{icml2024}

\newpage
\appendix
\onecolumn
\section{Appendix}
\label{sec:appendix}

\subsection{Dataset}
Data statistics of a full list of product categories are reported in Table \ref{tbl:data_full_cate}.
\begin{table} [htp]
\centering
\begin{tabular}{c|cccc}
\hline
\textbf{Category} & \textbf{products} & \textbf{labels} & \textbf{vocab.} & \textbf{complexity}\\
\hline
furn. & 2,174 & 23,210 & 3,278 & 1.75 $\pm$0.94\\
deco. & 1,054 & 10,481 & 2,664 & 1.77 $\pm$0.95\\
car & 181 & 2,925 & 741 & 1.90 $\pm$0.95\\
bed & 239 & 2,686 & 871 & 1.76 $\pm$0.96\\
outd. & 155 & 2,167 & 880 & 1.77 $\pm$0.94\\
stor. & 130 & 1,560 & 693 & 1.80 $\pm$0.93\\
acce. & 29 & 452 & 282 & 1.89 $\pm$0.98\\
home. & 23 & 382 & 278 & 1.78 $\pm$0.98\\
heat. & 21 & 377 & 184 & 1.69 $\pm$1.00\\
pers. & 23 & 344 & 227 & 1.93 $\pm$0.90\\
spor. & 23 & 324 & 161 & 1.74 $\pm$0.94\\
elec. & 17 & 310 & 193 & 1.90 $\pm$0.95\\
kitc. & 13 & 241 & 166 & 1.80 $\pm$0.92\\
offi. & 11 & 150 & 119 & 1.71 $\pm$0.98\\
all & 4,093 & 45,609 & 6,583 & 1.77 $\pm$0.94
\\\hline
\end{tabular}
\caption{Dataset counts for the products, labels and vocabulary size. 
Key: furn:furniture; deco:decor; car:car, bed:bed bath; outd:outdoor; stor:storage organization; acce:accessories; home:home improvement; heat:heating cooling; pers:personal electronic devices; spor:sports fitness; elec:home electronics; kitc:kitchen; offi:office}
\label{tbl:data_full_cate}
\end{table}


\subsection{Experiment Results}
The experiments results for a full list of product categories are presented in Table \ref{corr_human_whole_all_cate_1}, \ref{corr_human_whole_all_cate_2}, \ref{corr_human_high_all_cate_1} and \ref{corr_human_high_all_cate_2}.
\begin{table*}
\centering
\begin{tabular}{c|ccccccccc}
\hline
\textbf{Models} & \textbf{furn.} & \textbf{deco.} & \textbf{car} & \textbf{bed} & \textbf{outd.} & \textbf{stor.} & \textbf{acce.} & \textbf{home} & \textbf{heat.}\\
\hline
 $\Theta_v$ & .230 & .247 & .281 & .239 & .166 & .177 & .205 & .297 & .301\\
 $\Theta_s$ & .247 & .219 & .252 & .247 & .227 & .176 & .267 & .238 & .264 \\
$\Theta_r$ & .042 & .060 & .131 & .055 & .038 & .007 & .027 & -.109 & .159 \\
$\Theta_u$ & .175 & .177 & .207 & .230 & .174 & .131 & .136 & .247 & .290 \\
$\Theta_c$ & .180 & .178 & .171 & .208 & .163 & .121 & .129 & .116 & .218

\\\hline\hline
$\Theta_{v,s}$ & .258 & .252 & .266 & .259 & .221 & .185 & .279 & .295 & .305 \\
$\Theta_{v,s,u}$ & .267 & .270 & .272 & .276 & .236 & .204 & .277 & .376 & .267 \\
$\Theta_{v,s,u,c}$ & .273 & .276 & .316 & .294 & .242 & .199 & .275 & .294 & .347 \\\hline
\end{tabular}
\caption{Correlation with human ratings of complexity (the whole dataset)}
\label{corr_human_whole_all_cate_1}
\end{table*}

\begin{table*}
\centering
\begin{tabular}{c|cccccc}
\hline
\textbf{Models} & \textbf{pers.} & \textbf{spor.} & \textbf{elec.} & \textbf{kitc.} & \textbf{offi.} & \textbf{all}\\
\hline
 $\Theta_v$ & .236 &.139 & .164 & .242 & .224 & .225\\
 $\Theta_s$ & .291 & .128 & .126 & .250 & .258 & .228\\
$\Theta_r$ & .100 & -.029 & .041 & .089 & .073 & .049 \\
$\Theta_u$ & .173 & .090 & .061 & .155 & .109 & .168 \\
$\Theta_c$ & .160 & -.047 & .122 & .170 & .354 & .160

\\\hline\hline
$\Theta_{v,s}$ & .315 & .136 & .127 & .275 & .329 &.250\\
$\Theta_{v,s,u}$ & .316 & .153 & .129 & .278 & .329 & .261\\
$\Theta_{v,s,u,c}$ & .324 & .132 & .152 & .272 & .397 & .271
\\\hline
\end{tabular}
\caption{(Continued with Table \ref{corr_human_whole_all_cate_1}) Correlation with human ratings of complexity (the whole dataset)}
\label{corr_human_whole_all_cate_2}
\end{table*}



\begin{table*}
\centering
\begin{tabular}{c|ccccccccc}
\hline
\textbf{Models} & \textbf{furn.} & \textbf{deco.} & \textbf{car} & \textbf{bed} & \textbf{outd.} & \textbf{stor.} & \textbf{acce.} & \textbf{home} & \textbf{heat.} \\
\hline
$\Theta_v$ & .391 & .441 & \.471 & .513 & .222 & .381 & .128 & .499 & .425 \\
$\Theta_s$ & .425 & .375 & .497 & .520 & .178 & .390 & .471 & .343 & .384\\
$\Theta_u$ & .293 & .287 & .355 & .466 & .164 & .356 & .097 & .258 & .234\\
$\Theta_c$ & .277 & .278 & .190 & .404 & .101 & .337 & -.040 & .388 & -.058 
\\\hline\hline
$\Theta_{v,s}$ & .436 & .443 & .509 & .550 & .219 & .416 & .413 & .425 & .415 \\
$\Theta_{v,s,u}$ & .459 & .458 & .531 & .611	& .234 & .494 & .367 & .443 & .417 \\
$\Theta_{v,s,u,c}$ & .454 & .462 & .543 & .614	& .225 & .514 & .280 & .475 & .371

\\\hline
\end{tabular}
\caption{Correlation with human ratings of complexity (the subset with high rater agreement, correlation $> 0.75$)}
\label{corr_human_high_all_cate_1}
\end{table*}

\begin{table*}
\centering
\begin{tabular}{c|cccccc}
\hline
\textbf{Models} & \textbf{pers.} & \textbf{spor.} & \textbf{elec.} & \textbf{kitc.} & \textbf{offi.} & \textbf{all}\\
\hline
$\Theta_v$ & .447 & .657 & -.187 & .724 & .740 & .418\\
$\Theta_s$ & .447 & .515 & .004 & .630 & .621 & .414\\
$\Theta_u$ & -.787 & .501 & -.260 & .464 & .535 & .260 \\
$\Theta_c$ & .894 & -.240 & .017 & .425 & .774 & .267 
\\\hline\hline
$\Theta_{v,s}$ & .447 & .581 & -.25 & .677 & .784 & .433\\
$\Theta_{v,s,u}$ & .447 & .713 & -.242 & .662 & .784 
& .456 \\
$\Theta_{v,s,u,c}$ & .783 & .591 & -.180 & .652 & .827 & .472

\\\hline
\end{tabular}
\caption{(Continued with Table \ref{corr_human_high_all_cate_1}) Correlation with human ratings of complexity (the subset with high rater agreement, correlation $> 0.75$)}
\label{corr_human_high_all_cate_2}
\end{table*}

\end{document}